\documentclass[10pt,twocolumn,letterpaper]{article}
\usepackage{authblk}
\usepackage[pagenumbers]{arxiv} % To force page numbers, e.g. for an arXiv version

\usepackage{graphicx}
\usepackage{amsmath}
\usepackage{amssymb}
\usepackage{booktabs}
\usepackage[pdftex]{color}
\usepackage{enumerate}
\usepackage{multirow}

\usepackage[capitalize]{cleveref}
\crefname{section}{Sec.}{Secs.}
\Crefname{section}{Section}{Sections}
\Crefname{table}{Table}{Tables}
\crefname{table}{Tab.}{Tabs.}

\begin{document}

%\title{Human Body Segmentation using Airborne Ultrasound \\ via Collaborative Learning Variational Autoencoder}
\title{Invisible-to-Visible: Privacy-Aware Human Instance Segmentation \\ using Airborne Ultrasound via Collaborative Learning Variational Autoencoder}

\author[,1]{Risako Tanigawa \thanks{tanigawa.risako@jp.panasonic.com}}
\author[,1]{Yasunori Ishii \thanks{ishii.yasunori@jp.panasonic.com}}
\author[,1]{Kazuki Kozuka \thanks{kozuka.kazuki@jp.panasonic.com}}
\author[,2]{Takayoshi Yamashita \thanks{takayoshi@isc.chubu.ac.jp}}
\affil[1]{Technology Division, Panasonic Holdings Corporation}
\affil[2]{Machine Perception and Robotics Group, Chubu University}

\maketitle

%%%%%%%%% ABSTRACT
\begin{abstract}
In action understanding in indoor, we have to recognize human pose and action considering privacy.
Although camera images can be used for highly accurate human action recognition, camera images do not preserve privacy. %are feared to be a threat to privacy.
Therefore, we propose a new task for human instance segmentation from invisible information, especially airborne ultrasound, for action recognition.
To perform instance segmentation from invisible information, we first convert sound waves to reflected sound directional images (sound images).
Although the sound images can roughly identify the location of a person, the detailed shape is ambiguous.
To address this problem, we propose a collaborative learning variational autoencoder (CL-VAE) that simultaneously uses sound and RGB images during training.
In inference, it is possible to obtain instance segmentation results only from sound images.
As a result of performance verification, CL-VAE could estimate human instance segmentations more accurately than conventional variational autoencoder and some other models.
Since this method can obtain human segmentations individually, it could be applied to human action recognition tasks with privacy protection.
\end{abstract}

%%%%%%%%% BODY TEXT
%----------------------------------------------------------
% Introduction
%----------------------------------------------------------
\section{Introduction}
Recently, human action recognition has attracted wide attention~\cite{dhamsania2016survey, poppe2010survey} because of its applications, such as automated surveillance~\cite{BenMabrouk2018surveillance}, intelligent robots~\cite{Yanli2020HRI}, and behavior monitoring in homes~\cite{WEINLAND2011224, Gowsikhaa2014, Gruosso2021, alemdar2013aras, rai2021home}. % intelligent robotsの例
Several sensors, such as cameras~\cite{WEINLAND2011224}, electromagnetic waves~\cite{Xinyu2019radarHAR}, and wearable devices~\cite{abdel2021human, ramanujam2021human, wang2020wearable, cornacchia2016survey, zubair2016human}, have been employed to detect human activities. % cite
Among them, a camera is a human-friendly sensing device because we can intuitively understand the validity of recognition results by referring to the recorded videos.
Although camera-based human action recognition has been widely investigated and achieved high precision, there are environments where it is difficult to recognize human actions with cameras, \eg, dark environments.
In addition, camera images do not preserve privacy.%installing cameras in homes may cause privacy-related problems. % related to privacy.

\begin{figure}[t]
  \centering
  \includegraphics[width=0.9\linewidth]{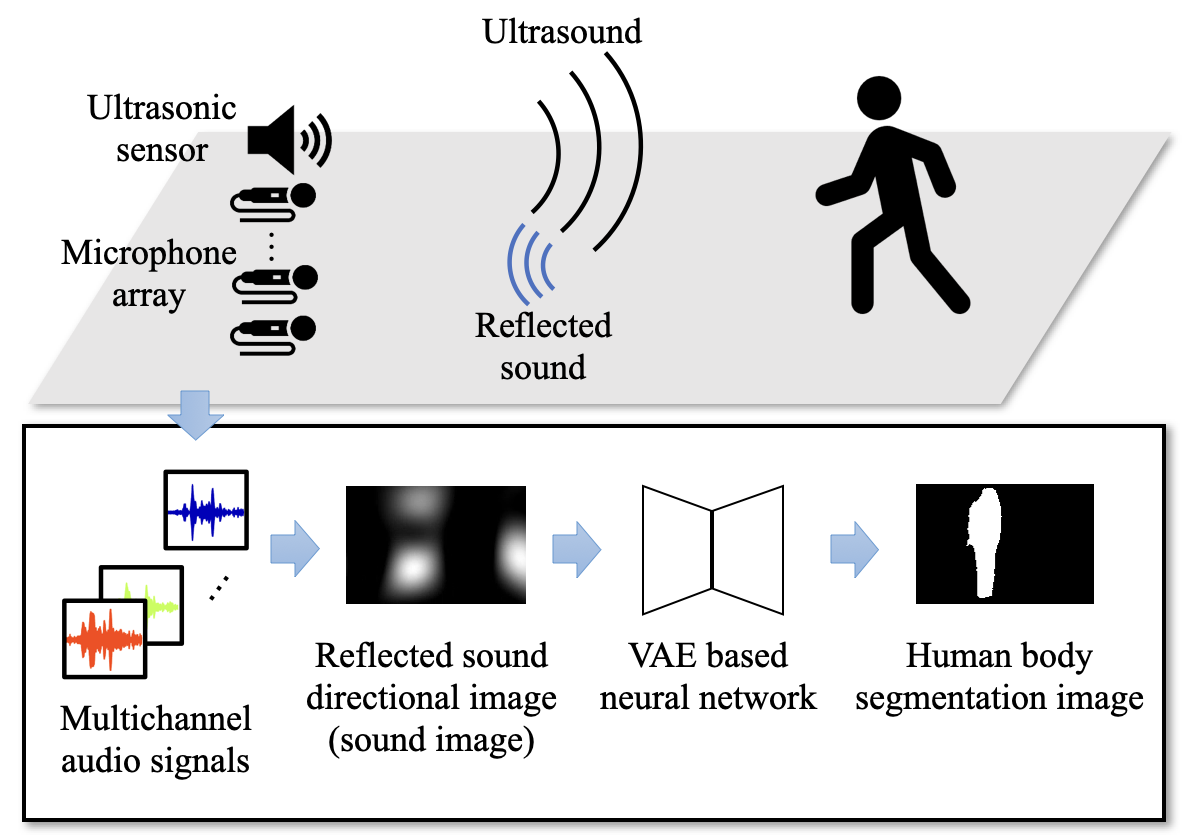}

   \caption{Concept of our work. The top row shows the setting of a sensing environment. The bottom row illustrates the concept of our method.}
   \label{fig:concept}
\end{figure}

Privacy-preserved human action recognition has been investigated using cameras other than RGB, such as event cameras~\cite{alonso2018evsegnet, Yi2020evmseg, liu2021event} and depth sensors~\cite{liang2015survey, Dinh2016depthseg, rahmani2017learning, wang2019comparative, Lee2019depthseg, Yin2020depthheight, wang20203dv, shaikh2021rgb}.
Even if we use an RGB camera, a method of recognizing human action by a shadow reflected on a white wall without directly capturing a person has been proposed~\cite{sharma2021learn}.
Further, cameraless approaches for human action recognition have been achieved by wearable sensors and phone sensors~\cite{abdel2021human, ramanujam2021human, wang2020wearable, cornacchia2016survey, zubair2016human}.
Although these methods can detect the part of motions, detecting actions that require whole-body information are difficult.
By contrast, whole-body segmentation has been achieved by electromagnetic waves, such as radars~\cite{Li2019radarsurvey, Zhao2018CVPR}, LiDARs~\cite{Maturana2015voxnet, Kunisada20181dcnn}, and WiFis~\cite{Wang2019survey, wang2019personinwifi, Kefayati2020wi2vi}.
These methods can detect persons without human images captured by cameras.
Although they are privacy-preserved sensing methods, it is difficult to capture the surrounding environment information.

Moreover, audio signals have been used to detect the surrounding environment information and humans without cameras~\cite{Vasudevan2020sounddepth, Irie2019sts, laput2018ubicoustics, moreira2020acoustic, sim2015acoustic, yatani2012bodyscope}.
Predicting depth maps and segmentations from audio signals have been proposed~\cite{Vasudevan2020sounddepth, Irie2019sts}.
These methods can generate images from invisible physical information and can be applied for human action recognition by analyzing human segmentation images.
Although sounding objects can be visualized by those methods, detecting nonsounding objects is difficult.
From the human recognition perspective, people who do not make sounds, such as not talking or walking, cannot be detected.

Using airborne ultrasound echoes is a possible technique to detect nonsounding people.
There are methods for detecting surrounding information by analyzing echoes~\cite{Liu2020acloc, Das2017gesture, murray2017bio, Hwang2019BatGnet, fu2015opportunities, parida2021image, xie2021hearfit}.
Although these methods can estimate the position of objects, methods that specialize in human recognition and estimate human instance segmentation have not been well investigated.
If the human instance segmentation can be estimated from echoes, it is possible to detect nonsounding people and it can be applied to estimate human action.
In addition, if it is possible to estimate using narrow-band ultrasounds without interfering with other bands, it can be expected to be used in combination with environmental sound analysis.
%cutting information on other bands,

In this study, we propose a method for estimating human instance segmentation from airborne ultrasound detected by multichannel microphones using a neural network.
The concept of our work is illustrated in \Cref{fig:concept}.
In the sensing section, an ultrasonic sensor at 62 kHz resonance, and 16 channels of microelectromechanical system (MEMS) microphone array are used.
%we used an ultrasonic sensor at 62 kHz resonance and 16 channels of microelectromechanical system (MEMS) microphone array.
The sound emitted from the ultrasonic sensor is reflected by a human body and then captured by microphones.
By analyzing the differences between multichannel audio signals, the reflected sound directional images (hereinafter, sound images) can be obtained.
Since the sound image represents the intensity of the reflected wave at each pixel, it has ambiguous shapes that are far from the human shape.
Therefore, we introduce a deep neural network to obtain human instance segmentation images from sound images.
The accuracy of conventional methods, such as U-Net~\cite{ronneberger2015unet} and Mask R-CNN~\cite{he2018mask}, reduces if the edges of objects between input and output are not similar, as mentioned in~\cite{Han2019unetedge}.
%Conventional segmentation methods such as U-Net~\cite{ronneberger2015unet} and Mask R-CNN~\cite{he2018mask} require that the edges of the objects between input and output should be similar as mentioned in~\cite{Han2019unetedge}.
However, the sound images have different edge positions from the segmentation images.
Thus, we propose a collaborative learning variational autoencoder (CL-VAE), which is based on variational autoencoder (VAE)~\cite{kingma2014vae}.
The CL-VAE learns latent variables from both segmentation and sound images in the training phase and estimates human instance segmentation images from sound images in the test phase.
Experiments showed that the human instance segmentation images could be generated from sound images.
To the best of our knowledge, this is the first work to estimate human instance segmentation images from airborne ultrasounds.

The main contributions of this work are as follows.
\begin{itemize}
      \item A human body sensing system using an airborne ultrasonic sensor and a microphone array for privacy-aware human instance segmentation. %combining an airborne ultrasonic sensor, a microphone array, as well as beamforming with fixed reflected-wave components subtracted.
      \item An architecture of CL-VAE, which learns latent variables from both segmentation and sound images in the training phase. %an encoder from both annotated and sound images.
      \item A loss function for CL-VAE, including a reconstruction error, Kullback-Leibler (KL) divergence, and mean square error (MSE) of parameters from the encoders between segmentation and sound images.
      \item Creation of a dataset for evaluating CL-VAE.
\end{itemize}

%----------------------------------------------------------
% Related works
%----------------------------------------------------------
\section{Related works}
%\subsection{Semantic segmentation}
% semantic segmentationの説明を入れる
% like: https://arxiv.org/pdf/2105.15203.pdf
%Semantic segmantation is pixel level image classification and variou deep learning methods, %such as Fully Convolutional Network (FCN)~\cite{long2015fcn}, U-Net %~\cite{ronneberger2015unet}, SegNet~\cite{badrinarayanan2016segnet}, and Pyramid Scene %Parsing network (PSPNet) \cite{zhao2017pspnet}, have been proposed.
%Since semantic segmentation classifies by pixel, it can categorize objects with more %complicated shapes than object detections.
%Therefore, semantic segmentation is used various fields such as autonomous driving, %robotic vision, and medical image processing.

%Though semantic segmentation typically categorizes pixels into classes from camera images, it can be regarded as a kind of visualization method when considering that segmentation images are generated from sensing devices other than cameras.

\subsection{Privacy-preserved human instance segmentation}
Several RGB camera-based human instance segmentation methods have been developped~\cite{WEINLAND2011224}.
Although the camera-based method has been well investigated and achieved high accuracy, privacy concerns should be considered for applications such as home surveillance.

Cameraless human instance segmentation methods have also been investigated.
Wang \etal~\cite{wang2019personinwifi} proposed a method of estimating human instance segmentation images, joint heatmaps, and part affinity fields using WiFi signals.
The channel state information~\cite{Halperin2011csi} was analyzed using the method.
They used three transmitting-receiving antenna pairs and thirty electromagnetic frequencies with five sequential samples.
The networks mainly comprise upsampling blocks, residual convolutional blocks, U-Net, and downsampling blocks.
Although this method achieved privacy-friendly fine-grained person perception with off-the-shelf WiFi antennas and regular household WiFi routers, it was not focused on environmental objects.

Alonso \etal~\cite{alonso2018evsegnet} proposed an event camera based semantic segmentation method.
The event camera senses information about pixels when the brightness changes, such as when the subject moves.
Since event cameras only capture the changes in intensities on a pixel-by-pixel basis, they do not capture personal information more clearly than the RGB cameras.
In~\cite{alonso2018evsegnet}, event information from event cameras was formed as 6-channel images. The first two channels were histograms of positive and negative events, whereas the other four channels were the mean and standard deviation of normalized timestamps at each pixel for the positive and negative events.
They showed that an Xception-based encoder-decoder architecture could learn semantic segmentation from the 6-channel information.
Although this method achieves semantic segmentation from privacy-preserved event cameras, it is difficult to detect people who are not moving.

Irie \etal~\cite{Irie2019sts} proposed a method that generates segmentation images from sounds.
They focused on the ability of humans to imagine the surrounding environmental scenes and developed the method.
They recorded sounds using four-channel microphone arrays.
To estimate segmentation images, they extracted Mel frequency cepstral coefficients and angular spectrum from sounds, which contain object types and their locations, respectively.
This method can estimate both human and environmental objects only from sounds.
However, in principle, it is difficult to estimate segmentation images for nonsounding people.
Therefore, we focused on airborne ultrasonic sensing to achieve the detection of nonsounding people.

\subsection{Airborne ultrasonic sensing}
Airborne ultrasonic sensors have been used in various industries, such as the automobile~\cite{Wang2014parking} and manufacturing industries~\cite{FANG2017uswood, CHIMENTI2014usmaterial, KAZYS2006usmultilayer}, to detect the distance from objects. % cite
Ultrasonic sensors emit short pulses at regular intervals.
If there are objects on the propagation path, the ultrasounds are reflected.
By analyzing the time differences between emitted and reflected sounds, the distance of objects can be detected~\cite{Carullo2001US}.
%This time-of-flight method can only detect distances.

In addition to the distances, the directions of objects can be detected when sounds are captured using microphone arrays~\cite{Moebus20073dus}.
Sound localization using beamforming algorithms has been developed. %cite, DAS, MUSIC
A common beamforming algorithm is the delay-and-sum (DAS) method~\cite{Perrot2021das}. %cite DAS
The DAS method can estimate the direction of sound sources by adding array microphone signals delayed by a given amount of time.
%These estimation methods have been applied for sound source separations. %他の応用例あれば追加
Considering a reflected position as a sound source, the positions of objects can also be estimated using ultrasonic sensors and microphone arrays.
%Ultrasonic sensing methods have been developed in robotic sensings to detect surrounding barriers.
Although these methods can detect the position of objects, it is difficult to obtain the actual shape of objects from echoes of a single pulse.

To detect the shape of objects, Hwang \etal~\cite{Hwang2019BatGnet} performed three-dimensional shape detections using wideband ultrasound and neural networks.
Although analysis of multiple frequencies can precisely detect the positions and shapes of objects, the speaker system for emitting wideband frequency sounds becomes large, and the amount of data increases due to the high sampling rate required to sense wideband ultrasound.
Therefore, we consider that obtaining segmentation images from the positional information on reflected objects analyzed by narrowband frequency ultrasound using a neural network.

%----------------------------------------------------------
% Proposed method
%----------------------------------------------------------
\section{Ultrasound sensing in proposed method}
Firstly, we describe the hardware setup of our ultrasound sensing system.
Then, we describe the preprocessing for converting ultrasound waves to sound images.

\subsection{Ultrasound sensing system}
\begin{figure}[t]
  \centering
  \includegraphics[width=0.9\linewidth]{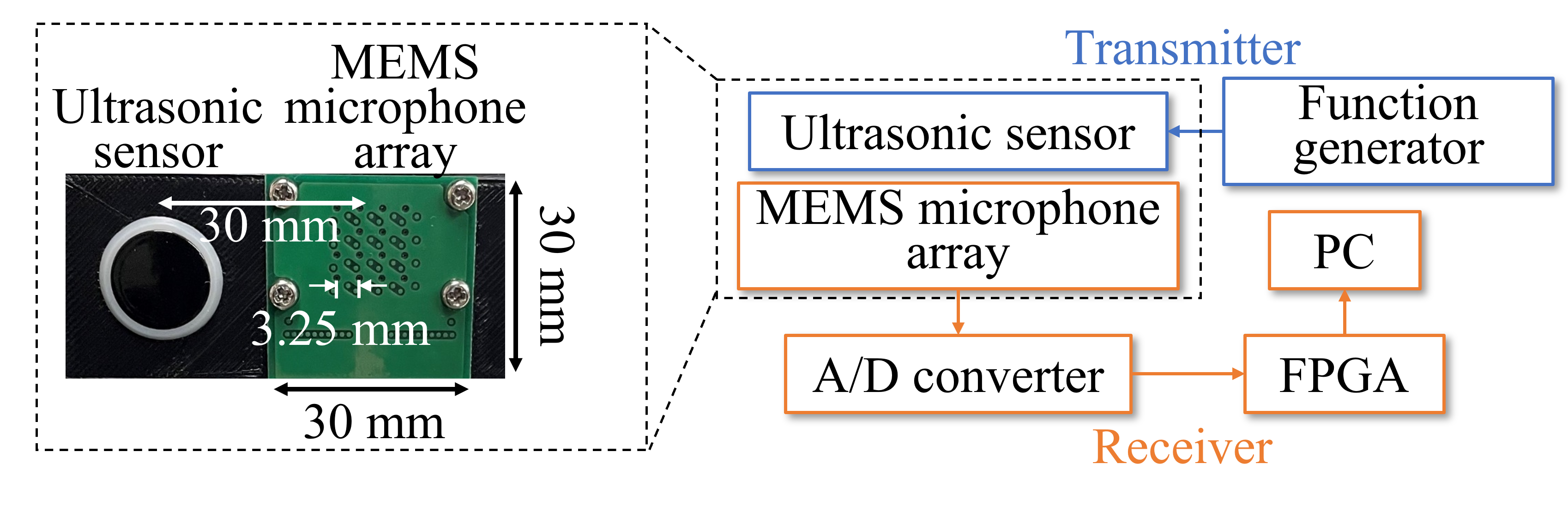}

   \caption{Hardware setup of ultrasonic wave transmitter and receiver system.}
   \label{fig:us_sensor}
\end{figure}

The hardware setup is shown in \Cref{fig:us_sensor}.
The transmitter comprised a function generator and an ultrasonic sensor at 62 kHz resonance.
%, which resonates around $62 \, \rm{kHz}$.
The ultrasonic sensor was driven with burst waves with $20$ cycles and $50 \, \rm{ms}$ intervals at $62 \, \rm{kHz}$ by the function generator.
The receiver comprised a MEMS microphone array, an analog-to-digital converter, a field-programmable gate array (FPGA), and a PC.
%A $4 \times 4$ grid MEMS microphone array was used.
%The $16$ MEMS microphones were mounted on a $30 \, \rm{mm^2}$ substrate at $3.25 \, \rm{mm}$ intervals.
A $4 \times 4$ grid MEMS microphone array, whose microphones were mounted on a $30 \, \rm{mm^2}$ substrate at $3.25 \, \rm{mm}$ intervals, was used.
The analog signals captured by the microphones were converted to digital signals and imported to the PC via the FPGA.
% 50msは往復17m（片道8.5m分の時間に相当するため人を検知するには充分な時間）
The distance between the microphone array and ultrasonic sensor was set to $30 \, \rm{mm}$.

\begin{figure*}[t]
  \centering
  \includegraphics[width=0.9\linewidth]{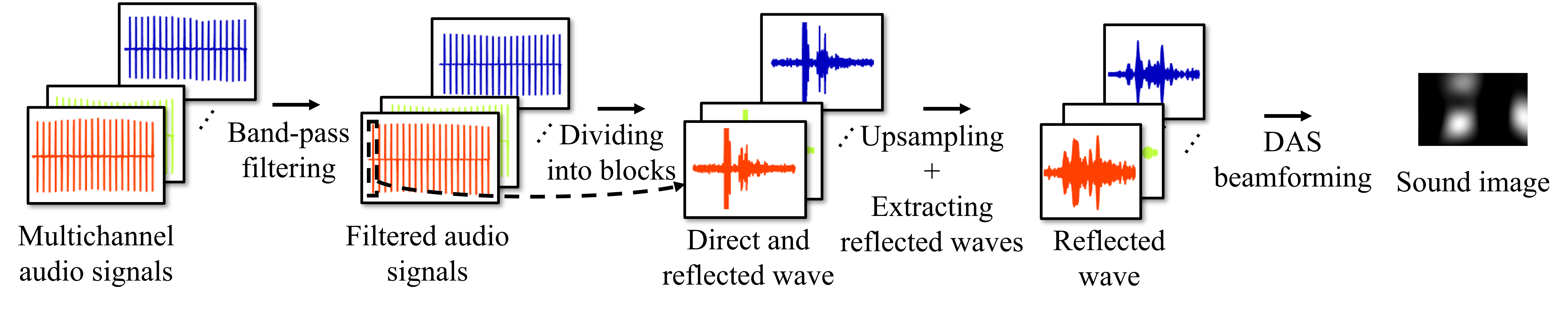}

   \caption{Diagram of preprocessing for audio signals. The multiple audio signals are filtered and then divided into single direct-reflected wave pairs. The upsampled reflected waves are extracted and DAS beamformed, and a sound image is created.}
   \label{fig:preprocess_arch}
\end{figure*}

\subsection{Data preprocessing}
%\Cref{fig:example_uswave} shows an example of the ultrasonic signals at one block which contains a direct sound generated from the ultrasonic sensor and reflected sounds from the human body.
%These waves have fluctuations with 62 kHz.
%The phase differences between microphones for direct and reflected sounds contain information about the direction of the ultrasonic sensor and the human body, respectively.
The diagram of the data preprocessing is shown in \Cref{fig:preprocess_arch}.
First, a band-pass filter with a center frequency of $62 \, \rm{kHz}$ and bandwidth of $10 \, \rm{kHz}$ was applied for audio signals captured by the $16$ microphones.
The filtered audio signals were divided into blocks including direct and reflected waves.
Then, we upsampled the soundwaves four times at each block because the $192 \, \rm{kHz}$ sampling rate was inadequate to represent sounds of $62 \, \rm{kHz}$.
Afterward, we produced sound images from reflected sounds via DAS beamforming.
The beamformed signal $y$ can be defined as
\begin{equation}
	y(t) = \sum_{m=1}^{M} x_m(t-\Delta_m),
\end{equation}
%$y(t) = \sum_{m=1}^{M} x_m(t-\Delta_m)$, 
where $t$ is the time, $M$ is the number of microphones, $x_m$ is signals received by the $m$-th microphone, and $\Delta_m$ is the time delay for the $m$-th microphone, which is determined by the speed of sound and the distance between the $m$-th microphone and observing points.
We calculated beamformed signals at the range of $\theta=-45$--$45$ degrees in the azimuthal direction and $\phi=-60$--$60$ degrees in polar direction.
Then we obtained the reflected directional heat maps.
To reduce the noises from reflected waves from objects, we calculated sound heat maps $H_{\rm{us}}$ by subtracting a reference map $H_{\rm{ref}}$ from reflected directional heat maps $H$ as
\begin{equation}
  H_{\rm{us}}(i,j) = H(i,j) - k H_{\rm{ref}}(i,j),
\end{equation}
where $(i,j)$ is the pixel of the heat maps, and $k$ is the coefficient, which is determined by
\begin{equation}
	k = \frac{H(i_{\rm{max}}, j_{\rm{max}})}{H_{\rm{ref}}(i_{\rm{max}}, j_{\rm{max}})},
\end{equation}
where $(i_{\rm{max}}, j_{\rm{max}})$ is the index of the maximum pixel of $H_{\rm{ref}}$.
Notably, the reference map was calcurated using the data without humans, and $H_{\rm{us}}$ was normalized as
\begin{equation}
	X_{\rm{us}}(i, j) =
	\begin{cases}
		0, & H_{\rm{us}}(i,j) < 0 \\
		\frac{H_{\rm{us}}(i,j)}{\max(H_{\rm{us}})}, & H_{\rm{us}}(i,j) \geq 0
	\end{cases}
  \label{eq:soundimage}
\end{equation}
when it was converted to sound images $X_{\rm{us}}$.

\begin{figure}[t]
  \centering
  \includegraphics[width=0.9\linewidth]{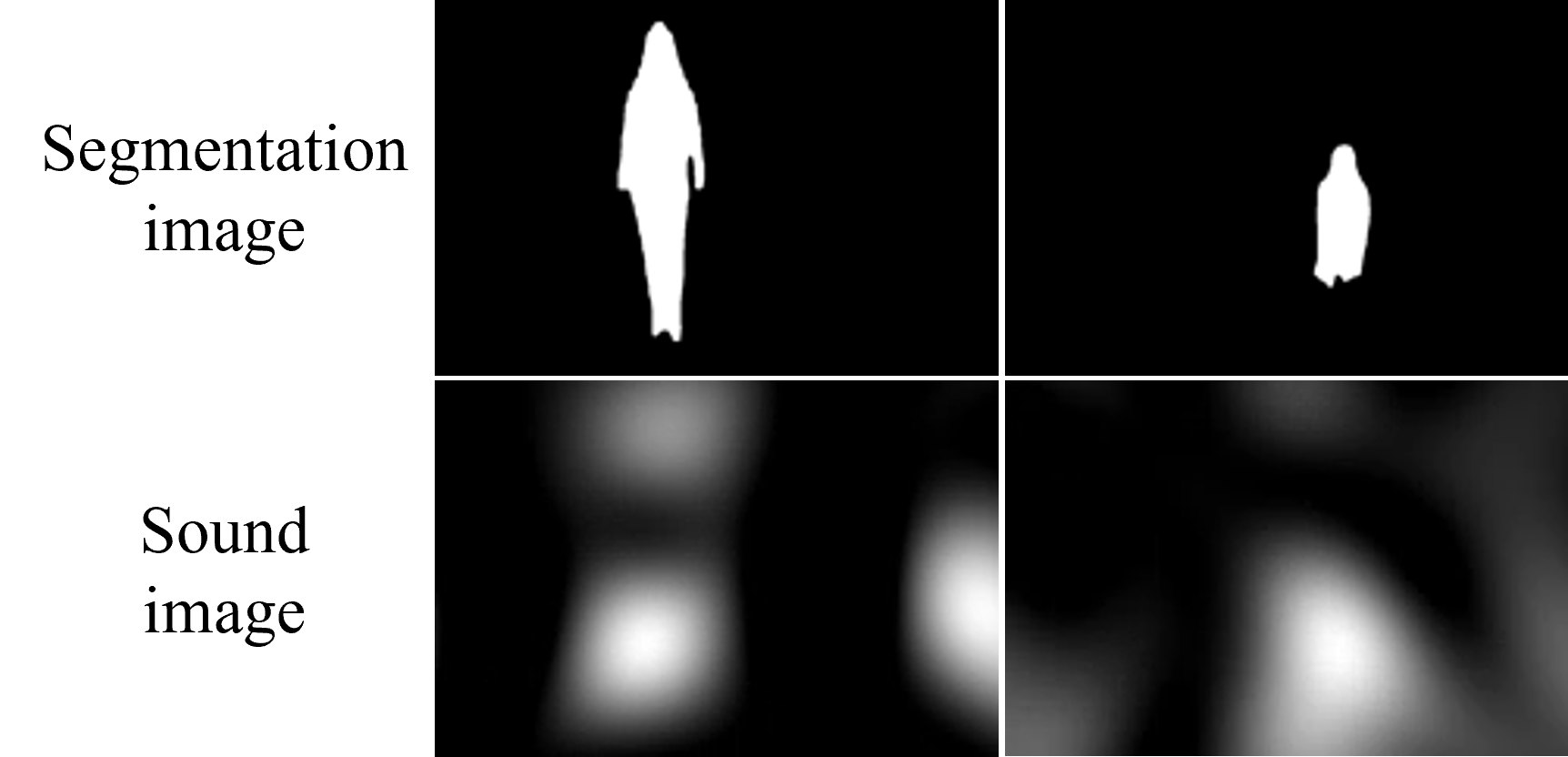}

   \caption{Example of sound images. The top row is the segmentation images genera by RGB images and the bottom row is the sound images.}
   \label{fig:soundimage}
\end{figure}

Examples of sound images are shown in \Cref{fig:soundimage}.
The segmentation images in the top row were annotated from RGB images.
The sound images, which represent the intensity of the reflected sound at each pixel, had ambiguous shapes at the positions corresponding to a person.
Furthermore, there were artifacts in the region with no person.

\section{Human instance segmentation via ultrasound}
We first describe an overview of the proposed network.
Second, we briefly explain VAE that is the basis of the proposed method.
Then, we describe the proposed network in detail.
Finally, we explain the loss function of our network.
\subsection{Overview}
As shown in \Cref{fig:soundimage}, the sound images do not have the shape of a person, and the edges are far different from those of the segmentation images.
In conventional segmentation methods, such as U-Net~\cite{ronneberger2015unet}, SegNet~\cite{badrinarayanan2016segnet}, and Mask R-CNN~\cite{he2018mask}, the accuracies are reduced where the edges of the objects between input and output are not similar.
%input/output images are required to have similar edges.
Han \etal~\cite{Han2019unetedge} stated that dividing regions from images with blurred edges was difficult.
Thus, we consider performing human instance segmentation based on probabilistic models, such as VAE, rather than deterministic models.
Kohl \etal~\cite{Simon2018unetvae} proposed a probabilistic U-Net combining VAE with U-Net.
Since this method fed values sampled from the latent space to the last layer of U-Net, it could not handle the ambiguity of edges.
Therefore, we propose CL-VAE, which introduces collaborative learning with the segmentation and sound images into VAE.
In CL-VAE, both segmentation and sound images were input to an encoder, and means and variances of each image were calculated at a training phase.
The loss includes the sum of the errors related to the distance between means and the distance between variances, in addition to the loss of VAE with the segmentation image.
By learning with the loss, the model learned to adapt the latent variable of the sound images to that of the segmentation images.
Consequently, both the segmentation and sound images are mapped to the common latent space; segmentation images can be estimated by generating images with a decoder from the latent variables sampled where the sound images are input into the model.
The details of the proposed method are described below.

\subsection{VAE~\cite{kingma2014vae}}
Since our network is based on VAE, we briefly explain it.
VAE is a generative model and has an encoder-decoder architecture.
The objective of VAE is to inference latent variables existing in datasets and generate data from the latent variables.
Let $X$ is a data and $z$ is a latent vector, the joint probability $p_{\theta} (X,z)$ is defined as $p_{\theta} (X,z) = p_{\theta} (X|z) p_{\theta} (z)$, where $p_{\theta} (X|z)$ is the conditional distribution of $X$ given $z$, $p_{\theta}(z)$ is the prior distribution of the latent vector $z$, and $\theta$ is the parameter of the generative model.
To inference the latent vector, the posterior $p_{\theta} (z|X)$ is calculated using Bayes's theorem:
%\begin{equation}
%  p_{\theta}(z|X) = \frac{p_{\theta}(X|z)p_{\theta}(z)}{p_{\theta}(X)},
%\end{equation}
$p_{\theta}(z|X) = p_{\theta}(X|z)p_{\theta}(z)/p_{\theta}(X)$, where $p_{\theta}(X)$ is the marginal liklihood, given by $p_{\theta}(X) = \int p_{\theta} (X|z) p_{\theta}(z) \, dz$.
%\begin{equation}
%  p_{\theta}(X) = \int p_{\theta} (X|z) p_{\theta}(z) \, dz.
%\end{equation}
The integral requires samplings from huge data.
Therefore, the approximate distribution of the posterior distribution $q_{\phi}(z|X)$ is introduced in VAE.
The approximate distribution $q_{\phi} (z|X)$ is learned as the encoder, and the generative model $p_{\theta}(X|z)$ is learned as the decoder.

The VAE learns the parameters of $\theta$ and $\phi$ by maximizing the marginal likelihood $p_{\theta}(X)$.
The objective function $\log p_{\theta}(X)$ can be written as %follows:
%\begin{equation}
%  \log p_{\theta}(X) = L(X,\theta,\phi) + D_{KL}(q_{\phi}(z|X)\parallel p_{\theta}(z|X)),
%\end{equation}
$\log p_{\theta}(X) = L(X,\theta,\phi) + D_{KL}(q_{\phi}(z|X)\parallel p_{\theta}(z|X))$, 
where $L(X,\theta,\phi)$ is the evidence lower bound (ELBO) and $D_{KL}$ is the KL divergence.
Since the KL divergence is nonnegative, maximizing the marginal likelihood can be considered as maximizing the ELBO.
%The ELBO can also be written as 
%\begin{align}
%  L(X,\theta,\phi) &= E_{q_{\phi}(z|X)} [\log p_{\theta}(X,z)-\log q_{\phi}(z|X)] \nonumber \\
%  &= \log p_{\theta}(X) - D_{KL}(q_{\phi}(z|X) \parallel p_{\theta}(z|X)).
%\end{align}
To realize the backpropagation in the optimization of the ELBO, the reparametrization trick is introduced.
The sampling of $z$ is alternatively performed by another random variable as $z = \mu + \epsilon \sigma$, where $\mu$ and $\sigma$ are the mean and variance of the posterior distribution $q_{\phi}(z|X)$, respectively, and $\epsilon \sim \mathcal{N}(0,I)$.
The loss function of the VAE comprises a reconstruction error and a regularization term and can be written as
%\begin{equation}
%  L_{\rm{VAE}} = L_{\rm{RE}} + D_{\rm{KL}}(q_{\phi}(z|X)\parallel p_{\theta}(z|X)),
%\end{equation}
$L_{\rm{VAE}} = L_{\rm{RE}} + D_{\rm{KL}}(q_{\phi}(z|X)\parallel p_{\theta}(z|X))$,
where $L_{\rm{RE}}$ is the reconstruction error.

\subsection{CL-VAE}
\begin{figure}[t]
  \centering
  \includegraphics[width=0.9\linewidth]{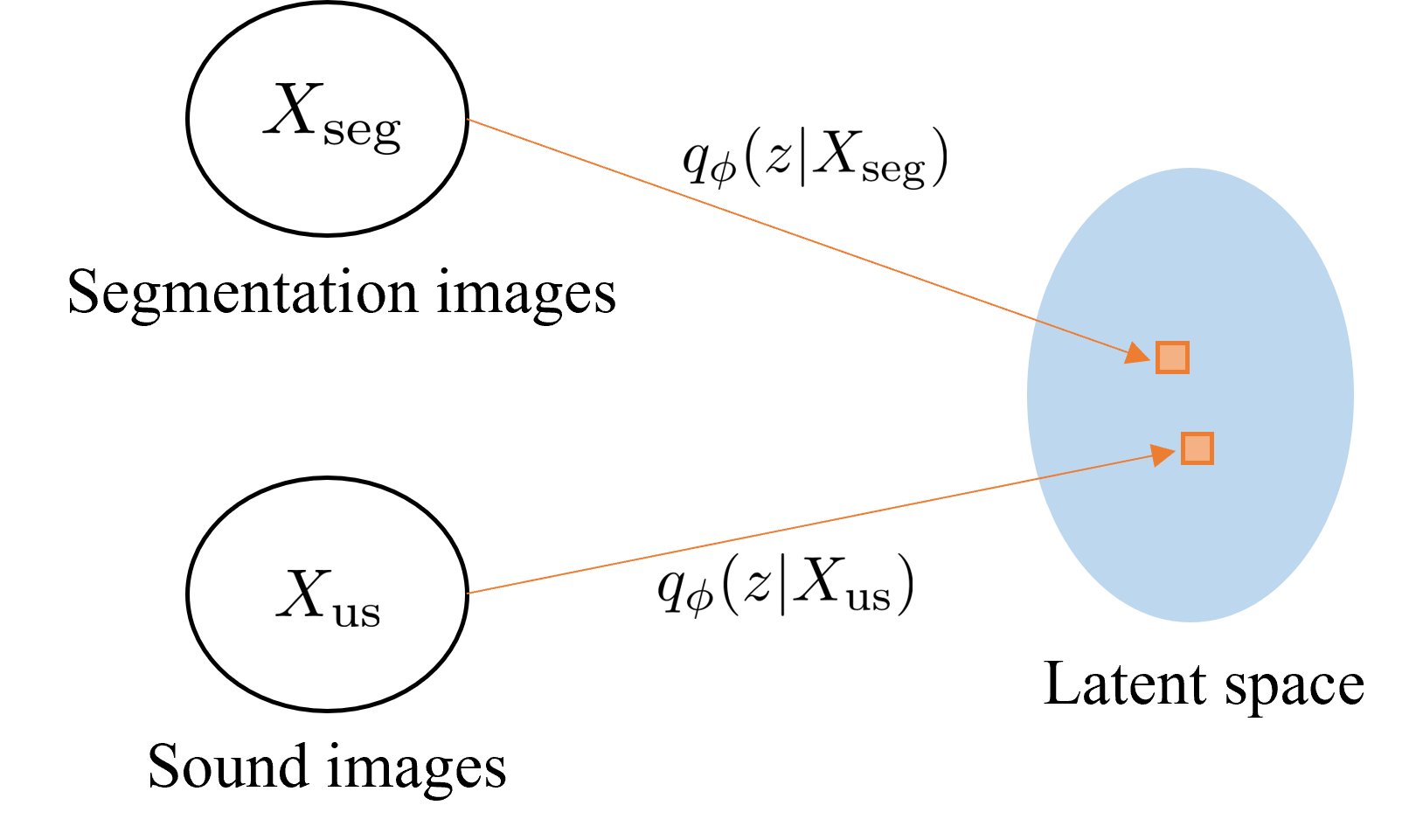}

   \caption{Concept of CL-VAE.}
   \label{fig:clvae_concept}
\end{figure}

The VAE can reconstruct images by obtaining the parameters of the probability distribution of the input dataset.
However, sound images differ from human instance segmentation images.
To overcome the problem, we considered that both the segmentation images $X_{\rm{seg}}$ and sound images $X_{\rm{us}}$ as input to the encoder and trained them to bring the distributions $q_{\phi} (z|X_{\rm{seg}})$ and $q_{\phi}(z|X_{\rm{us}})$ closer.
The encoder $q_{\phi} (z|X) \sim N(X; \mu_{\phi}, \sigma_{\phi})$ is the distribution of the latent space to the input $X$, represented by the means $\mu_{\phi}$ and the variances $\sigma_{\phi}$.
The decoder $p_{\theta}(X|z)$ generates plausible images for the input by sampling latent variables $z$ from the latent space.
Hence, if the encoder is trained to map to the same point in the latent space regardless of whether segmentation or sound images are input, the images generated from the latent variables obtained by inputting the sound images become close to the segmentation images (see \Cref{fig:clvae_concept}).
Since the encoder is represented by a Gaussian distribution with means and variances, it is possible to map to a latent space common to both images by matching the mean and variance of the segmentation and sound images.

\begin{figure}[t]
  \centering
  \includegraphics[width=0.9\linewidth]{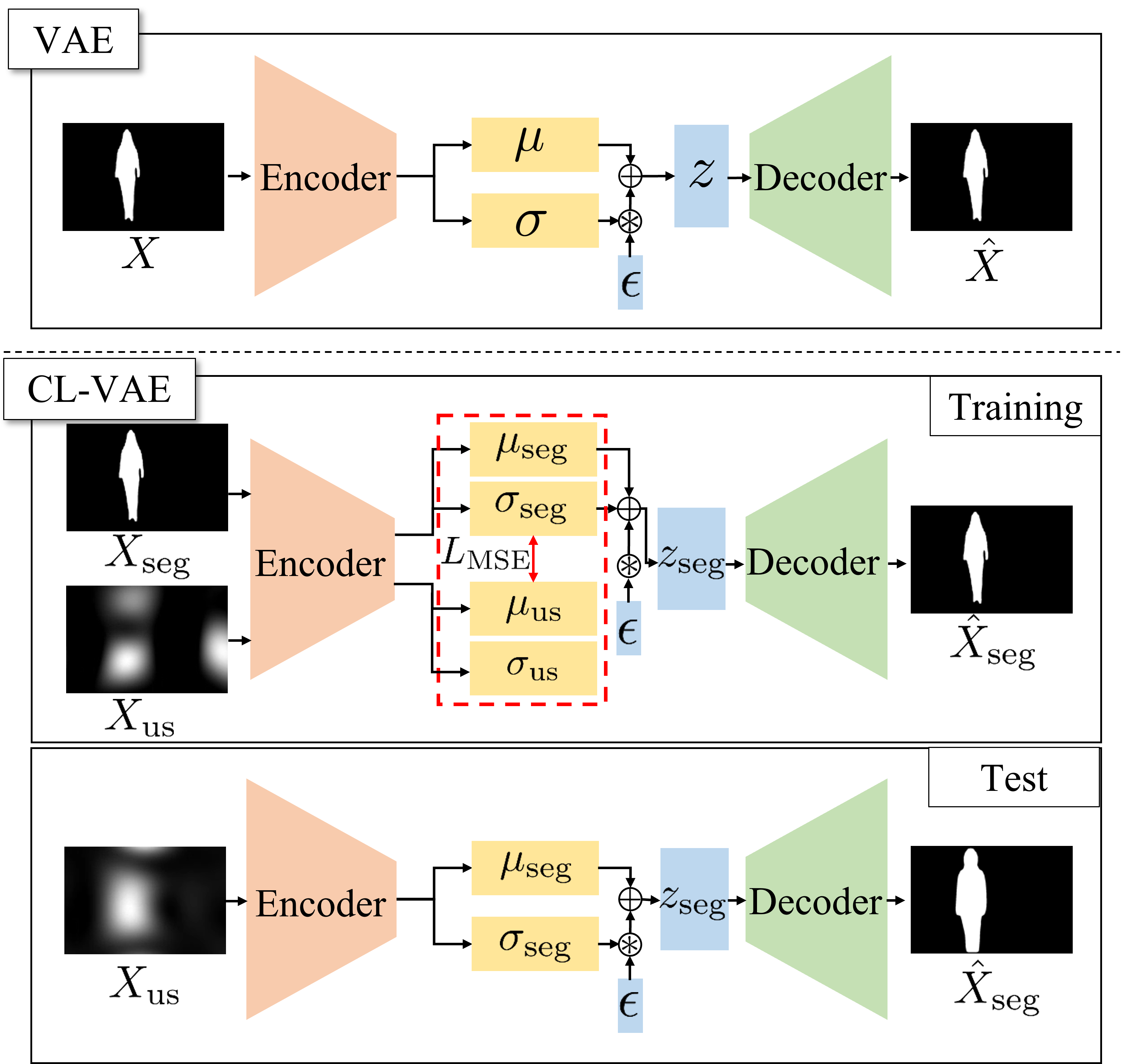}

   \caption{Schematic of VAE and our network. The top row depicts the network of VAE and the bottom row depicts CL-VAE.}
   \label{fig:network}
\end{figure}

The diagram of our network is shown in \Cref{fig:network}.
At the training phase, both the segmentation images $X_{\rm{seg}}$ and sound images $X_{\rm{us}}$ are input to the encoder.
Then, each of them is encoded, and the means $\mu_{\rm{seg}}, \mu_{\rm{us}}$ and variances $\sigma_{\rm{seg}}, \sigma_{\rm{us}}$ are estimated from the posterior distributions $q_{\phi} (z|X_{\rm{seg}})$ and $q_{\phi}(z|X_{\rm{us}})$, respectively.
The parameters $\mu_{\rm{us}}$ and $\sigma_{\rm{us}}$ are trained to approximate the parameters $\mu_{\rm{seg}}$ and $\sigma_{\rm{seg}}$ by comparing their values.
%To train the decoder model $p_{\theta} (z|X_{\rm{seg}})$, the latent vector of the segmentation images are reparametarized as $z_{\rm{seg}} = \mu_{\rm{seg}} + \epsilon \sigma_{\rm{seg}}$.
At the testing phase, the sound images are input to the encoder, %and the means and variances are reparameterized to the latent vector.
%Then, the latent vector is input to the decoder, 
and the segmentation images are obtained.

\subsection{Loss functions}
To train the distribution $q_{\phi} (z|X_{\rm{us}})$ to become close to the distribution $q_{\phi}(z|X_{\rm{seg}})$, we introduce MSEs of the means and variances of the distributions $q_{\phi}(z|X_{\rm{seg}})$ and $q_{\phi} (z|X_{\rm{us}})$.
The loss is defined as %function of our model is
\begin{align}
	L = \; &\alpha \{ L_{\rm{RE}} + D_{\rm{KL}}(q_{\phi}(z|X_{\rm{seg}})\parallel p_{\theta}(z|X_{\rm{seg}}))\} \nonumber \\
  &+ (1-\alpha)L_{\rm{MSE}}.
\end{align}
The first term is the sum of the reconstruction loss and KL divergence, which are the same as in the conventional VAE.
%the reconstruction loss and $D_{\rm{KL}}$ is the KL divergence, 
The second term is the MSE of means and variances.
%the loss of a difference in latent variable.
%The first term is the loss of the VAE and the second term is the loss of the encoding block.
To adjust the scales of the VAE loss and the MSE loss, the coefficient $\alpha$ is introduced.
The reconstruction error $L_{\rm{RE}}$ is calculated as
\begin{align}
	L_{\rm{RE}} = \frac{1}{N} &\sum_{n=1}^{N} (-x_{\rm{seg}\it{,n}} \log \hat{x}_{\rm{seg}\it{,n}} \nonumber \\
  &- (1-x_{\rm{seg}\it{,n}}) \log(1-\hat{x}_{\rm{seg}\it{,n}})),
\end{align}
where $x_{\rm{seg}}$ is the value of input segmentation images, $\hat{x}_{\rm{seg}}$ is the value of reconstructed images, $N$ is the dimension of input/output images, and $n$ is the index of the dimension.
The KL divergence $D_{\rm{KL}}$ is calculated as
\begin{align}
	&D_{\rm{KL}}(q_{\phi}(z|X_{\rm{seg}})\parallel p_{\theta}(z|X_{\rm{seg}})) \nonumber \\
  &= -\frac{1}{2} \sum_{d=1}^{D} (1+\log(\sigma_{\rm{seg}\it{,d}}^2)-\mu_{\rm{seg}\it{,d}}^2-\sigma_{\rm{seg}\it{,d}}^2),
\end{align}
where $D$ is the dimension of the latent variables $z$, and $d$ is the index of the dimension.
The $L_{\rm{MSE}}$ is the MSE between $\mu_{\rm{seg}}, \sigma_{\rm{seg}}$ and $\mu_{\rm{us}}, \sigma_{\rm{us}}$ calculated as
\begin{equation}
	L_{\rm{MSE}} = \frac{1}{D} \sum_{d=1}^{D} (\mu_{\rm{us}\it{,d}} - \mu_{\rm{seg}\it{,d}})^2 + \frac{1}{D} \sum_{d=1}^{D} (\sigma_{\rm{us}\it{,d}} - \sigma_{\rm{seg}\it{,d}})^2.
\end{equation}

%----------------------------------------------------------
% Experiment and result
%----------------------------------------------------------

\section{Experments}
First, we describe the experimental setup.
Then, the results of the experiments are explained.
\subsection{Experimental setup}
\paragraph{Inplementation details}
The batch size was $128$ and the initial learning rate was $0.001$.
We used an Adam optimizer with $\beta_1 = 0.9 , \beta_2 = 0.999$ in training.
We trained the network for $40$ epochs.
The coefficient of the loss function $\alpha$ was set to $0.0001$, which was experimentally determined by confirming the scales of $L_{\rm{RE}}$, $D_{\rm{KL}}$, and $L_{\rm{MSE}}$ in advance.
Our network was implemented by PyTorch.

\begin{figure*}[t]
  \centering
  %\fbox{\rule{0pt}{2in} \rule{1.0\linewidth}{0pt}}
  \includegraphics[width=0.9\linewidth]{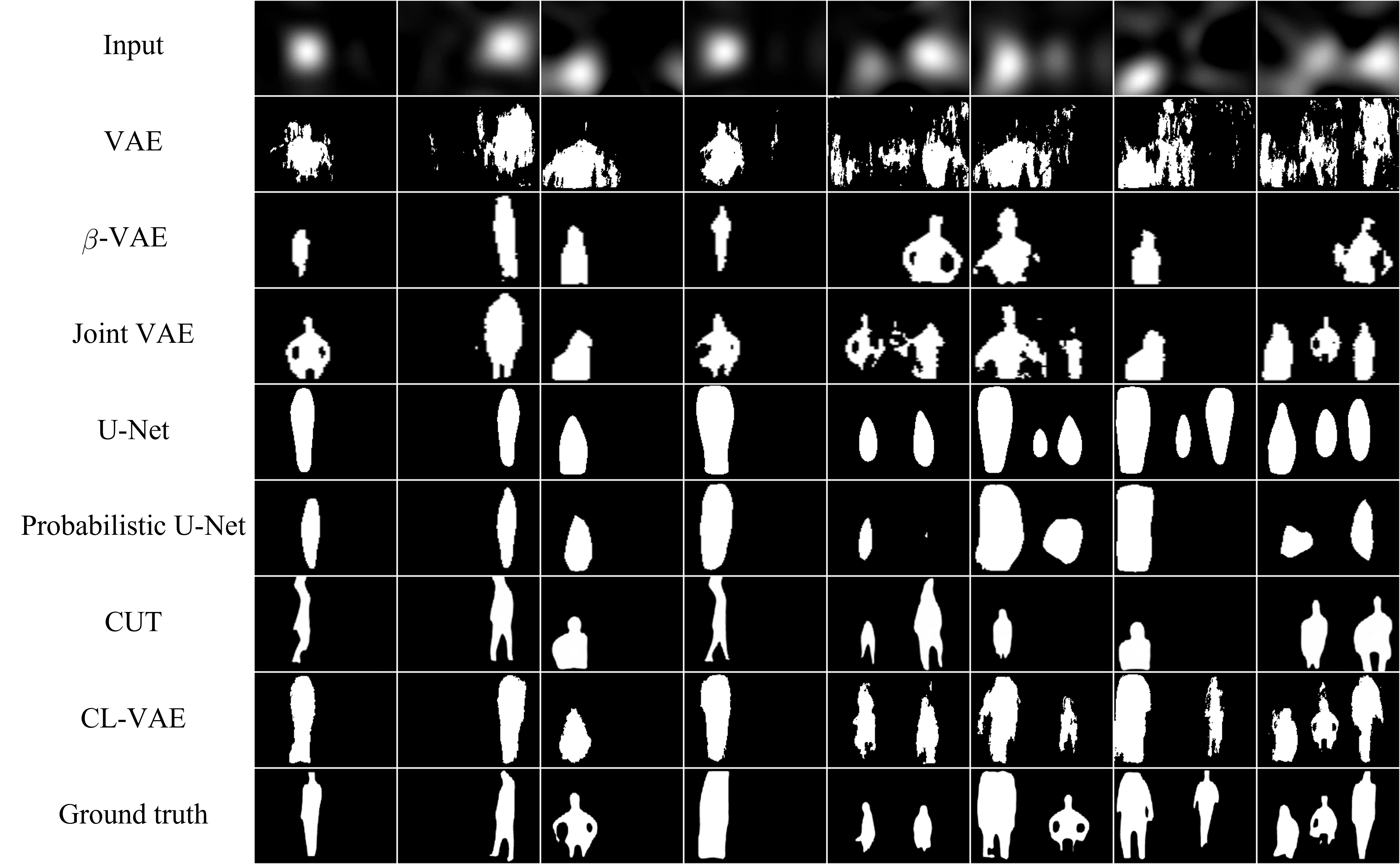}

   \caption{Qualitative results. The top row is input, the second to seventh rows are estimated segmentation images by conventional methods, the eighth row is estimated segmentation images by CL-VAE, and the bottom row is ground truth.}
   \label{fig:predicted}
\end{figure*}

\begin{table*}[t]
  \centering
  \caption{Quantitative results of human instance segmentation.}
  \vspace{-1mm}
  \scalebox{1.0}{
   \begin{tabular}{cccc}
     \toprule
     \multirow{2}{*}{Method} & \multicolumn{2}{c}{Training}  & \multirow{2}{*}{mIoU} \\ 
     %\cline{2-3}
      & Input & Label & \\
     \midrule
     %\cline{1-4}
      VAE~\cite{kingma2014vae} & Segmentation & - & 0.293 \\ 
      $\beta$-VAE~\cite{Higgins2017betaVAE} & Segmentation & - & 0.325 \\
      JointVAE~\cite{dupont2018jointVAE} & Segmentation & - & 0.282 \\
      U-Net \cite{ronneberger2015unet} & Sound & Segmentation & 0.560 \\
      Probabilistic U-Net~\cite{Simon2018unetvae} & Sound & Segmentation & 0.549 \\
      CUT~\cite{park2020contrastive} & Sound and segmentation & - & 0.292 \\
      \midrule
      CL-VAE & Sound and segmentation & - & 0.532 \\
     \bottomrule
   \end{tabular}
  }
  \label{tab:mIoU}
\end{table*}

%\begin{table}[t]
%  \centering
%  \caption{mIoU comparison with other generative models.}
%  \begin{tabular}{lclclc}
%    \toprule
%    model & mIoU \\
%    \midrule
%    Probabilistic U-Net~\cite{Simon2018unetvae} & 0.653 \\
%    $\beta$-VAE~\cite{Higgins2017betaVAE} & 0.331 \\
%    JointVAE~\cite{dupont2018jointVAE} &  0.305 \\
    %CAR-GAN \cite{duan2019cascade} & \\
%    CUT~\cite{park2020contrastive} &  0.324\\
%    \midrule
%    Ours & 0.543 \\
%    \bottomrule
%  \end{tabular}
%  \label{tab:mIoU_generative}
%\end{table}

\vspace{-0.5cm}
\paragraph{Datasets}
We created a dataset for our experiment because no datasets use airborne ultrasound to detect the human body so far.
%\color{red}
Written consent was obtained from participants in the data acquisition.
%The dataset will not be publicly released because the participants did not agree for their data to be shared publicly.
%\color{black}
We captured the ultrasounds at $192 \, \rm{kHz}$ sampling from $16$ channel microphones and videos at $30$ frames per second (fps) from the RGB camera (a built-in camera of Let's Note, CF-SV7, Panasonic), which was located $35 \, \rm{mm}$ under the microphone array, for $10 \, \rm{s}$.
The resolution was $180 \times 120$, and the videos were used for creating segmentation images used in a training phase.
The data were extracted at $10$ fps because the time interval of the ultrasonic sound generation was $20$ bursts per second and the frame rate of the video was $30$ fps.
We produced segmentation images using Mask R-CNN~\cite{he2018mask}.
We used the dataset that people, who were located from $1$ to $3 \, \rm{m}$ away from the sensing devices, performed continuous motions such as standing, sitting, walking, and running in scenes. 
Three people performed in two scenes.
The total number of images was $19,982$; $80$\% and $20$\% were used for training and testing, respectively.

\vspace{-0.5cm}
\paragraph{Evaluation metrics}
We evaluated the performance of the model using a mean intersection-over-union (mIoU). %求め方？
To calculate the mIoU of the output images from the VAE-based methods, including CL-VAE, the sigmoid function was used for the last layer of the decoders, and the decoded images were binarized at a threshold of $0.5$.

\begin{figure*}[t]
  \centering
  %\fbox{\rule{0pt}{2in} \rule{1.0\linewidth}{0pt}}
  \includegraphics[width=0.92\linewidth]{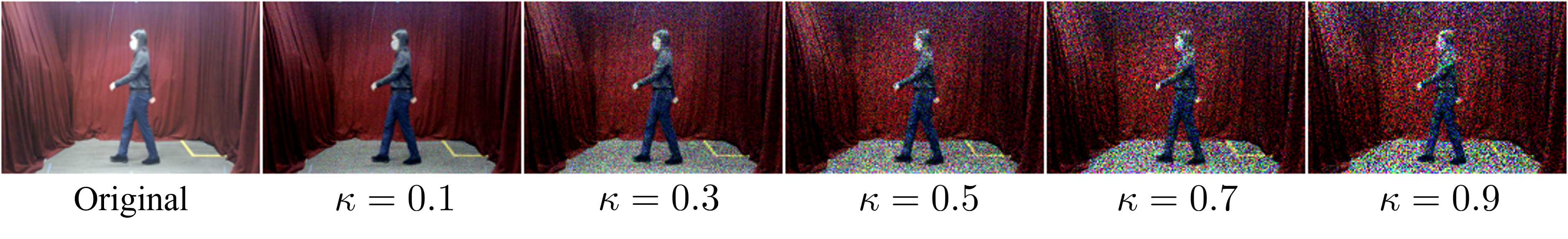}

   \caption{RGB and noisy images for evaluations using Mask R-CNN.}
   \label{fig:cam_noise}
\end{figure*}

\begin{table*}[t]
  \centering
  \caption{Comparison of mIoU with camera-based approach. Mask R-CNN pretrained by COCO dataset, which had less noise, was used for the evaluation.}
  \vspace{-1mm}
  \scalebox{0.9}{
   \begin{tabular}{c|c|cccccccccc|c}
     \toprule
      & RGB images & \multicolumn{10}{|c|}{Noisy images} & CL-VAE \\
     \midrule
     $\kappa$ & - & 0.1 & 0.2 & 0.3 & 0.4 & 0.5 & 0.6 & 0.7 & 0.8 & 0.9 & 1.0 & - \\
     PSNR & - & 14.7 & 14.4 & 14.2 & 13.7 & 13.3 & 12.9 & 12.4 & 12.1 & 11.7 & 11.4 & - \\
     \midrule
     mIoU & 0.764 & 0.731 & 0.689 & 0.619 & 0.545 & 0.472 & 0.384 & 0.310 & 0.240 & 0.202 & 0.174 & 0.532 \\
     \bottomrule
   \end{tabular}
  }
  \label{tab:mIoU_camera}
\end{table*}

\subsection{Experimental results}
To evaluate the performance, we first describe the performance of the proposed CL-VAE and compare it with other methods.
Then, we describe the comparison with the segmentation of RGB images.
Finally, we explain the performance of CL-VAE against environmental changes.
%the evaluation of the performance of CL-VAE against environmental changes.

\vspace{-1mm}
\subsubsection{Performance of CL-VAE}
We first evaluate the performance of the proposed CL-VAE.
To confirm the validity of learning with both sound and segmentation images, we compared our model with VAE, $\beta$-VAE~\cite{Higgins2017betaVAE}, Joint VAE~\cite{dupont2018jointVAE}, U-Net, Probabilistic U-Net~\cite{Simon2018unetvae}, and contrastive unpaired translation (CUT)~\cite{park2020contrastive}.
VAE, $\beta$-VAE, Joint VAE were trained by segmentation images and inferenced by sound images.
U-Net and Probabilistic U-Net were trained by the sound images with segmentation label images and inferenced by sound images.
CUT and CL-VAE were trained by sound and segmentation images and inferenced by sound images.
The image sizes were $64 \times 64$ in $\beta$-VAE and Joint VAE, $128 \times 128$ in Probabilistic U-Net, $256 \times 256$ in CUT, and $180 \times 120$ in VAE, U-Net, and CL-VAE.
The mIoUs were calculated with images resized to $180 \times 120$.

\Cref{tab:mIoU} and \Cref{fig:predicted} illustrate the quantitative and qualitative evaluations, respectively.
The mIoU of our model was higher than those of other models, except for U-Net and Probabilistic U-Net.
Although the mIoUs of U-Net and Probabilistic U-Net were higher than CL-VAE, the shapes of people were ambiguous in the estimated images compared with those of CL-VAE as shown in \cref{fig:predicted}.
%This result showed that learning with sound images improved the precision due to the closer latent variables between the sound and segmentation images.
%This result shows that learning with sound images improved the precision thanks to adapting the distribution between the sound images and the segmentation images.
The images estimated by VAE were noisy and the shapes of people were not clearly estimated.%influenced by the artifacts in the region with no person which occurred when the sound images were generated.
On the other hand, the shapes of people of CL-VAE could be estimated more clearly than that of VAE. %without being affected by the artifacts in CL-VAE.
% In addition, the little complicated shape of the human body can be estimated in ours such as arms and legs.
Since the latent variables obtained from the sound and segmentation images by CL-VAE were close, it was possible to estimate using the edge information on the segmentation images even when the sound images were input.
%We thought that these are because CL-VAE acquired the correspondence between the distributions of the sound images and the annotated images in the training phase, and the model became robust to the artifacts. %and possible to estimate complex shapes not found in the input images.
%In $\beta$-VAE and Joint VAE, the positions of people in the estimated images were highly affected by the input image shapes.
In $\beta$-VAE, only a single person segmentation was estimated under conditions of multiple people, and the shapes tended to be different from the ground truth.
Although the segmentation of multiple people was estimated by Joint VAE, the shapes were different from the ground truth.
Though CUT segmentation images estimated the human shapes more clearly than our model, some people were not estimated especially in multiple people images, and had different shapes from the segmentation images.
%some images containing artifacts in the input images estimated distorted human shapes.

%\begin{figure*}[t]
%  \centering
  %\fbox{\rule{0pt}{2in} \rule{1.0\linewidth}{0pt}}
%  \includegraphics[width=0.9\linewidth]{fig/Fig10_result_gen.png}

%   \caption{Qualitative results of ours and other generative models. The top row is the input images. From second to sixth rows are the estiimated images by Probabilistic U-Net~\cite{Simon2018unetvae}, $\beta$-VAE~\cite{Higgins2017betaVAE}, Joint VAE~\cite{dupont2018jointVAE}, CUT~\cite{park2020contrastive}, and ours, respectively. The bottom row is the Ground truth.}
%   \label{fig:predicted_generative}
%\end{figure*}

%To visualize the learning process of the CL-VAE, we performed reconstructions of the segmentation and sound images at 1, 5, 10, 15, and 20 epochs.
%The example images were shown in \Cref{fig:decoded}.
%The reconstruction images $\hat{X}_{\rm{seg}}$ and $\hat{X}_{\rm{us}}$ were calculated from the segmentation images $X_{\rm{seg}}$ and sound images $X_{\rm{us}}$, respectively.
%The same weights at each epoch were used in the encoder and the decoder.
%The segmentation reconstructed images $\hat{X}_{\rm{seg}}$ approaches the ground truth as the number of epochs increases.
%Along with that, the shapes of the person in the sound images also become similar to the segmentation images.

\subsubsection{Gaps with camera-based approaches}
\vspace{-2mm}
Further, we compared our method with camera-based approaches.
To confirm the instance segmentation performance from RGB images, the mIoU of segmentation images estimated by Mask R-CNN was evaluated. 
The ground truth images were manual annotations of $500$ images selected from the test images.
The manual annotation was performed using LabelMe~\cite{labelme2016}.
The RGB images are required to be more sensitive to improve visibility in dark environments.
In that case, the noises contained in the images increase, which affects the estimation accuracy of the segmentation.
Besides, ultrasounds are unaffected by the changes in brightness.
Therefore, to compare the performance in a dark environment, we created noisy images that simulate images with high ISO sensitivity.
The mIoUs of the model trained by RGB images and tested by noisy images were evaluated.
Notably, the noisy images were generated as follows.
First, the brightness of the image $I$ was decreased, and the darkened image $I_{\rm{dark}}$ was generated as
\begin{align}
	&I_{\rm{dark}} = \frac{I}{255} - 0.2, \\
	&I_{\rm{dark}}(i,j) = 
	\begin{cases}
		0, & I_{\rm{dark}}(i,j) < 0 \\
		I_{\rm{dark}}(i,j), & I_{\rm{dark}}(i,j) \geq 0.
	\end{cases}
\end{align}
Then, noisy image $I_{\rm{noise}}$ was created by adding Gaussian noises $N$ with the mean $0$ and variance $0.5$ to the darkened image $I_{\rm{dark}}$.
\begin{equation}
	I_{\rm{noise}}(i,j) = I_{\rm{dark}}(i,j) + \kappa \sqrt{I_{\rm{dark}}(i,j)} N(i,j),
\end{equation}
where $\kappa$ is the coefficient to adjust the peak signal-to-noise ratio (PSNR) of the noisy images.
Examples of noisy images are shown in \Cref{fig:cam_noise}.
Mask R-CNN model pretrained by the COCO dataset~\cite{lin2015microsoft} was used for the estimation.

The mIoUs of the RGB and noisy images are shown in \Cref{tab:mIoU_camera}.
The mIoUs with noisy images decreased with a decrease in PSNRs.
On the other hand, our model could estimate segmentation images stably, even in dark environments, since our model was unaffected by brightness.
There are quantitative gaps between camera images and CL-VAE.
Considering that Mask R-CNN was trained on a large dataset, our model could reduce the gaps by increasing the number of data. %and improving the quality of the annotation images.

% \begin{figure*}[t]
%   \centering
%   \fbox{\rule{0pt}{2in} \rule{1.0\linewidth}{0pt}}
%   %\includegraphics[width=1.0\linewidth]{fig/Fig5_result.png}
%
%    \caption{Qualitative results.}
%    \label{fig:predicted_camera}
% \end{figure*}

\subsubsection{Performance of CL-VAE against environmental changes}
To evaluate the robustness of environmental changes, we conducted inferences by the test data captured by the following conditions.
\begin{enumerate}[(a)]
  \setlength{\parskip}{0cm}
  \setlength{\itemsep}{0cm}
  \item A person not included in the training dataset. We used a mannequin to simulate that condition.
  \item A change of a sensor position. Although we captured the training data in front of people, we captured the data from an angle in this condition.
  \item A change of ultrasound frequency from $62$ to $40 \, \rm{kHz}$. To match this condition, the analysis frequency band was changed in the preprocessing.
  \item A change of the distance between the wall and the sensor from $3$ to $1.5 \, \rm{m}$. To match this condition, the analysis section was limited to the range of $1.5 \, \rm{m}$ when generating the sound images.
\end{enumerate}

\begin{table}[t]
  \centering
  \caption{mIoU in various conditions.}
  \vspace{-1mm}
  \scalebox{1.0}{
   \begin{tabular}{lc}
     \toprule
      Condition & mIoU \\
     \midrule
      (a) Person & 0.357 \\
      (b) Sensor position & 0.215 \\
      (c) Ultrasound frequency & 0.206 \\
      (d) Distance of the wall & 0.489 \\
     \bottomrule
   \end{tabular}
  }
  \label{tab:mIoU_environment}
\end{table}

The mIoUs of these conditions are shown in \Cref{tab:mIoU_environment}.
The mIoUs decreased under all conditions. %as shown in \Cref{tab:mIoU_environment}.
%In the evaluation of (a) and (b), the mIoUs decreased due to the influence of the reflection conditions from the person and the difference in the shape of the person. 
%In the evaluation of (c), the mIoU decreased due to the influence of the effect of the difference in sound images derived from the difference in reflected waveforms.
%The mIoU of (d) did not significantly decrease due to adjusting the analysis section of sound image generation.
%\color{red}
In the evaluation of (a), it was assumed that the mIoU decreased because the shape of the mannequin was different from that of the training dataset and the reflected waves differed because of the surface material.
In the evaluation of (b), (c), and (d), it was assumed that the mIoUs decreased because the sound images were different from that of the dataset due to the difference in the reflection angle, the frequency, and the distance of the wall.
%\color{black}
%In the evaluation of (c), it was assumed that the mIoU decreased due to the influence of the effect of the difference in sound images derived from the difference in reflected waveforms.
To address the limitations obtained from these results, we will increase the variety of data and improve the sound image generation process.
%The accuracies will be improved by increasing the variety of data and improving the sound image generation process.

%----------------------------------------------------------
% Conclusion
%----------------------------------------------------------
\section{Conclusions}
We proposed privacy-aware human instance segmentation from airborne ultrasonic using CL-VAE.
Our approach can produce human instance segmentation images by learning the latent space shared by both segmentation and sound images for training.
%using both segmentation and sound images for training as well as learns the latent space shared by both images.
This approach can be applied to detect human actions %without capturing camera images.
in situations where consideration for privacy is required, such as home surveillance, because the sound/segmentation images cannot be reconstructed to RGB images.
%Therefore, it can be applied to situations where consideration for privacy is required, such as home surveillance.
To improve the accuracy in unknown environments, we will increase the variety of data and improve the sound image generation process in the future.

%%%%%%%%% REFERENCES
% 候補
% person segmentation for surveillance by RGB image https://link.springer.com/article/10.1007%2Fs10462-012-9341-3
% person segmentation for surveillance by infrared image https://link.springer.com/chapter/10.1007/978-3-642-13022-9_35
% automotive sensor review https://ieeexplore.ieee.org/abstract/document/8585340
%\clearpage
{\small
\bibliographystyle{ieee_fullname}
\bibliography{egbib}
}

\end{document}